\documentclass[sn-mathphys,Numbered]{sn-jnl}

\usepackage{graphicx}%
\usepackage{multirow}%
\usepackage{amsmath,amssymb,amsfonts}%
\usepackage{amsthm}%
\usepackage{mathrsfs}%
\usepackage[title]{appendix}%
\usepackage{xcolor}%
\usepackage{textcomp}%
\usepackage{manyfoot}%
\usepackage{booktabs}%
\usepackage{algorithm}%
\usepackage{algorithmicx}%
\usepackage{algpseudocode}%
\usepackage{listings}%

\theoremstyle{thmstyleone}%
\theoremstyle{thmstyletwo}%

\theoremstyle{thmstylethree}%

\begin{document}

\title{Beyond Tides and Time: Machine Learning's Triumph in Water Quality Forecasting}


\author[1,2]{\fnm{Yinpu Li}}\email{yinpulee@gmail.com}
\equalcont{These authors contributed equally to this work.}
\author*[1,2]{\fnm{Siqi Mao} }\email{mden17g@gmail.com}
\equalcont{These authors contributed equally to this work.}

\author[1,2]{\fnm{Yaping Yuan}}\email{yypyuan@gmail.com}
\equalcont{These authors contributed equally to this work.}

\author[1,2]{\fnm{Ziren Wang} }\email{zaynw1013@gmail.com}
\equalcont{These authors contributed equally to this work.}

\author[3]{\fnm{Yixin Kang}}\email{kangyixin512@gmail.com}

\author[3]{\fnm{Yuanxin Yao} }\email{yaoyuanxin2017@gmail.com}

\affil[1]{\orgdiv{Department of Statistics}, \orgname{Florida State University}, \orgaddress{\street{600 W College Ave}, \city{Tallahassee}, \postcode{32306}, \state{FL}, \country{USA}}}

\affil*[2]{\orgdiv{Department of Statistics}, \orgname{Florida State University}, \orgaddress{\street{600 W College Ave}, \city{Tallahassee}, \postcode{32306}, \state{FL}, \country{USA}}}

\affil[3]{\orgdiv{Department of Mathematics and Statistics},
\orgname{University of Massachusetts Amherst}, \orgaddress{\street{710 N. Pleasant Street}, \city{Amherst}, \postcode{01003}, \state{MA}, \country{USA}}}

\affil[4]{\orgdiv{Department of Statistics}, \orgname{Rice University}, \orgaddress{\street{6100 Main St}, \city{Houston}, \postcode{77005}, \state{TX}, \country{USA}}}

\affil[5]{\orgdiv{Department of Statistics}, \orgname{Florida State University}, \orgaddress{\street{600 W College Ave}, \city{Tallahassee}, \postcode{32306}, \state{FL}, \country{USA}}}
  
\affil[6]{\orgdiv{Department of Statistics}, \orgname{Florida State University}, \orgaddress{\street{600 W College Ave}, \city{Tallahassee}, \postcode{32306}, \state{FL}, \country{USA}}}


\abstract{
Water resources are essential for sustaining human livelihoods and environmental well-being. Accurate water quality prediction plays a pivotal role in effective resource management and pollution mitigation. In this study, we assess the effectiveness of five distinct predictive models—linear regression, Random Forest, XGBoost, LightGBM, and MLP neural network—in forecasting pH values within the geographical context of Georgia, USA. Notably, LightGBM emerges as the top-performing model, achieving the highest average precision. Our analysis underscores the supremacy of tree-based models in addressing regression challenges, while revealing the sensitivity of MLP neural networks to feature scaling. Intriguingly, our findings shed light on a counterintuitive discovery: machine learning models, which do not explicitly account for time dependencies and spatial considerations, outperform spatial-temporal models. This unexpected superiority of machine learning models challenges conventional assumptions and highlights their potential for practical applications in water quality prediction. Our research aims to establish a robust predictive pipeline accessible to both data science experts and those without domain-specific knowledge. In essence, we present a novel perspective on achieving high prediction accuracy and interpretability in data science methodologies. Through this study, we redefine the boundaries of water quality forecasting, emphasizing the significance of data-driven approaches over traditional spatial-temporal models. Our findings offer valuable insights into the evolving landscape of water resource management and environmental protection.}




\maketitle

\section{Introduction}\label{sec1}

In today's era, the significance of our water resources' quality is unparalleled, influencing every aspect of human existence and the natural world. Water, a fundamental necessity for survival, holds implications that extend well beyond quenching mere thirst. The presence of polluted water sources and deteriorated aquatic ecosystems has given rise to a multitude of problems encompassing public health, ecological balance, economic factors, and societal fairness. The proactive monitoring and projection of water quality play a pivotal role in protecting aquatic ecosystems such as rivers, lakes, and oceans. Deteriorated water quality can lead to the devastation of habitats, adverse effects on aquatic organisms, and disruption of the overall ecological equilibrium.

Among the critical parameters for evaluating water quality, the water's pH stands out as one of the most critical factors. It quantifies the level of acidity or alkalinity within the water. Water possessing a pH value of 11 or higher has the potential to induce irritation in the eyes, skin, and mucous membranes, underscoring the importance of pH assessment in water quality investigations. As highlighted by Geetha et al. (2016) \cite{geetha2016internet}, the Internet of Things (IoT) is also contributing to advancements in water quality monitoring. Consequently, the prediction of water quality, particularly in relation to pH levels, has become increasingly imperative in recent times.

Recognizing the significance of maintaining water quality, the need extends beyond mere monitoring to encompass proactive prediction. This proactive approach guarantees timely public alerts regarding potential contamination, subsequently averting the associated health risks and economic losses. There are many traditional water quality prediction methods, such as multiple linear regression \cite{rajaee2015forecasting} and auto-regressive integrated moving average (ARIMA) \cite{araghinejad2013data}. However, the linear nature of multiple linear regression poses a limitation in detecting nonlinear relationships among water quality parameters \cite{nourani2016self}. Similarly, the primary drawback of ARIMA lies in its underlying assumption of linearity \cite{zare2011forecasting}. During the process of model identification, the time series data must undergo scrutiny to determine their stationarity, a crucial aspect in constructing the ARIMA model. Notably, conventional methodologies struggle to effectively capture the non-linear \cite{huo2013using} and non-stationary \cite{chang2016estimating} characteristics inherent in water quality due to their intricate and complex nature.

In recent years, machine learning approaches have been widely applied to multiple domains and achieved gratifying results (e.g., \cite{chen2023tests, li2022adaptive, henrique2019literature}). When it comes to estimating water quality using machine learning, Lu et al. \cite{lu2020hybrid} applied two hybrid decision tree-based water quality machine learning models:  extreme gradient boosting (XGBoost) and random forest (RF), proposed to obtain more accurate short-term water quality prediction results, by using the water resources of Gales Creek site in Tualatin River. Huang et al. (2019) \cite{huang2019integrated} established a prediction system for urban estuary water quality and used the gradient boosting machine model to fill and predict the flow. Wang et al. (2022) \cite{wang2022prediction} applied several machine learning models—multiple linear regression, artificial neural networks, random forest, and extreme gradient boosting (XGBoost)—were developed to predict $\text{NH4}^{+}-N$ in the Xiaoqing River estuary, China.  The shapely additive explanations method \cite{NIPS2017_8a20a862} was used to interpret the XGBoost model and discover the influence of the upper reaches of the river on the estuary. In the research \cite{li2022interpretable}, Li et al. (2022) evaluated five tree-based models, namely classification tree, random forest, CatBoost, XGBoost, and LightGBM, and employ a state-of-the-art explanation method SHAP to explain the models. 

 As the volume of data continues its relentless expansion, traditional approaches are proving inadequate to cope with the demands of researchers. With the advent of increased computing power, data-driven models like artificial neural networks (ANNs) have undergone substantial improvements. These models excel at capturing the inherent functional relationships inherent within water quality data, as evidenced by examples in Zhang et al. (1998) \cite{zhang1998forecasting}.  Even in situations where articulating intricate data relationships proves challenging, ANNs have proven their effectiveness. Moreover, ANNs require fewer initial assumptions \cite{anmala2015gis} while delivering heightened precision \cite{li2019water} compared to established techniques. Furthermore, Singh et al. (2009) \cite{singh2009artificial} utilized the ANN model to predict the water quality of the Gomti River in India, showcasing the versatility of these models. García-Alba et al., 2019 \cite{garcia2019artificial} employed an ANN-based model to predict estuary bathing water quality, integrating laboratory analysis, machine learning, and numerical simulation for real-time water quality management.
Peng et al. (2019) \cite{peng2019development} proposed a framework for real-time prediction of daily water quality, successfully applying it to Lake Chaohu in China, thereby improving predictions for parameters such as dissolved oxygen and total phosphorus. 

The primary aim of this inquiry is to offer enlightenment to individuals who possess a strong background in data science but may be less acquainted with the realm of environmental research. We present a comprehensive framework for the application of data science knowledge and methodologies, facilitating their conversion into tangible applications across diverse research domains. This encompasses areas such as water quality prediction, enabling the practical utilization of these skills in various contexts. 

In this study, we test the proposed framework by applying the dataset used in \cite{zhao2019spatial}. This input data consists of daily water quality samples from 37 sites, providing measurements related to pH values in Georgia, USA. The input features consist of 11 common indices including the volume of dissolved oxygen, temperature, and specific conductance. The proposed framework is examined by forecasting water quality in terms of the “power of hydrogen (pH)” value based on the input data.

The remaining sections of the paper are structured as follows: In Section 2, we delve into the acquired data and the prediction methodology. This section extensively elucidates the various models employed in our study, encompassing Linear Regression, XGBoost, LightGBM, Random Forest, and a Multiple-layer Perceptron Neural Network. It also encompasses a thorough description of the evaluation metrics employed and provides a comprehensive overview of the entire implementation process. Moreover, this section entails an in-depth performance comparison across different models, followed by a detailed analysis. Additionally, we include a SHAP (SHapley Additive exPlanations) analysis within this section. Section 3 succinctly elucidates why our machine learning models outperform the original model, which accounts for both time dependency and spatial factors. Lastly, Section 4 encapsulates the main conclusions drawn from the study and outlines potential avenues for future research.

\section{Material and methods}\label{sec2}

\subsection{Data}
The daily monitoring data for 37 sites in Georgia, USA from 2016 to 2018 was collected by the United States Geological Survey\footnote{USGS: https://www.usgs.gov/. Accessed Feb, 2018}. The dataset presented in the original paper \cite{zhao2019spatial}, and it consists of 11 features with measurements related to PH values. The training set includes $37 \times 423$ elements and the test set includes $37 \times 282$ elements.   
More details are presented in Table \ref{data}.

\begin{table}[h]
\caption{Data Description.}\label{data}
\begin{tabular*}{\textwidth}{@{\extracolsep\fill}ccc}
\toprule%
Variable Name   & Size          & Description   \\ 
\midrule
features        & 1$\times$11   & A list of water indices to measure. \\
location\_ids   & 37$\times$1   & IDs of the water stations. \\
$X_{te}$        & 1$\times$282  & \begin{tabular}[c]{@{}c@{}}Test set input data: water indices for 282 contiguous dates \\ until 2018-01-01; each element is a 37 11 matrix: \\37 spatial locations by 11 features.\end{tabular} \\
$X_{tr}$        & 1$\times$423  &\begin{tabular}[c]{@{}c@{}}Training set input data: water indices for 423 contiguous\\  dates from 2016-01-28; each element is a 37*11 matrix: \\37 spatial locations by 11 features.\end{tabular}  \\

$Y_{te}$        & 37$\times$282 & \begin{tabular}[c]{@{}c@{}} Test set output data: water quality for 37 locations \\in 282 contiguous dates until 2018-01-01.\end{tabular}                                                                                             \\
$Y_{tr}$        & 37$\times$423 &\begin{tabular}[c]{@{}c@{}} Training set output data: water quality for 37 locations \\in 423 contiguous dates from 2016-01-28.\end{tabular}                                                                                    \\
location\_group & 1$\times$3    & \begin{tabular}[c]{@{}c@{}}The groups of water stations, each group forms \\a connected spatial network (i.e., water system).  \end{tabular}                                                                                             \\ 
\botrule
\end{tabular*}
\end{table}

\subsection{Data Preparation \& Feature Engineering}

To deal with the spatio-temporal data structure, the raw data need to be reorganized. 
\subsubsection{Data Stacking}

The original dataset comprises $N_{tr} = 423$ and $N_{te} = 282$ consecutive dates within the training and test datasets, respectively. Each date corresponds to a data matrix of dimensions $K \times p = 37 \times 11$, where $K$ denotes spatial locations, and $p$ represents the number of features.

We amalgamated the training and test datasets chronologically and spatially, resulting in a total of $N^{tr} \times p = (N_{tr} \times K) \times p = 15,651\times 11$ data points in the training set, and $N^{te} \times p = (N_{te} \times K) \times p = 10,434\times 11$ data points in the test set. In addition, the spatio-temporal features (Date, Location ID, Month, Week, Weekday, Season) are merged to the training and testing data.

\subsubsection{Feature Engineering}
\begin{enumerate}
    \item \textbf{Time Features:} Temporal Decomposition: We extract the time-based feature and created the ``Date" column, and extended it to ``Year", ``Month", ``Day", ``Day of the Week", etc.

    \item \textbf{Spatial Features:} Location Encoding: When stacking the matrices, we added a column ``Location\_ID" and applied one-hot encoding for feature processing. This allows us to include spatial information in the model.

\end{enumerate}

\subsection{Exploratory Data Analysis}

The features have long name in the raw data. For convenience purposes, we mapped the variable names to their simplified version. More details are presented in Table \ref{feature_name}.

\begin{table}[h]
\caption{Summary Statistics for Numerical features.}
\label{feature_name}
\begin{tabular*}{\textwidth}{@{\extracolsep\fill}lc}
    \toprule
Original feature Name  &  Simplified Name\\ 
    \midrule
\begin{tabular}[l]{@{}l@{}}Specific conductance, water, unfiltered, microsiemens per centimeter \\at 25 degrees Celsius (Maximum)\end{tabular} & X1 \\ 

pH, water, unfiltered, field, standard units (Maximum) & X2\\ 

pH, water, unfiltered, field, standard units (Minimum) & X3 \\ 

\begin{tabular}[l]{@{}l@{}}Specific conductance, water, unfiltered, microsiemens per centimeter \\at 25 degrees Celsius (Minimum) \end{tabular} & X4\\ 

\begin{tabular}[l]{@{}l@{}}Specific conductance, water, unfiltered, microsiemens per centimeter \\at 25 degrees Celsius (Mean) \end{tabular} & X5 \\ 

Dissolved oxygen, water, unfiltered, milligrams per liter (Maximum) & X6\\ 

Dissolved oxygen, water, unfiltered, milligrams per liter (Mean) & X7\\ 

Dissolved oxygen, water, unfiltered, milligrams per liter (Minimum) & X8\\ 

Temperature, water, degrees Celsius (Mean) & X9\\ 

Temperature, water, degrees Celsius (Minimum) & X10 \\ 

Temperature, water, degrees Celsius (Maximum) & X11 \\ 

pH, water, unfiltered, field, standard units (Median) &  Y\\

\bottomrule
  \end{tabular*}
\end{table}%

For performing exploratory data analysis, we provide summary statistics in Table \ref{Summary_Statistics}. All features are bounded between $0$ and $1$. In addition, no missing values are identified in the entire data set (both training and testing data). 

\begin{table}[h]
\caption{Summary Statistics for Numerical features (Training data).}\label{Summary_Statistics}
\begin{tabular}{@{}lrrrrrrrrrrrr@{}}
\toprule
{Statistics} &       X1 &       X2 &       X3 &       X4 &       X5 &       X6 &       X7 &       X8 &       X9 &      X10 &      X11 & Y \\ 
\midrule
Mean  &     0.07 &     0.89 &     0.03 &     0.04 &     0.57 &     0.86 &     0.60 &     0.58 &     0.56 &     0.54 &     0.55  &         0.66\\
SD    &     0.16 &     0.03 &     0.12 &     0.13 &     0.12 &     0.03 &     0.15 &     0.17 &     0.20 &     0.21 &     0.19  &         0.03 \\
Min   &     0.00 &     0.58 &     0.00 &     0.00 &     0.12 &     0.72 &     0.08 &     0.03 &     0.06 &     0.03 &     0.09 &         0.57 \\
25\%   &     0.00 &     0.87 &     0.00 &     0.00 &     0.49 &     0.84 &     0.52 &     0.48 &     0.39 &     0.37 &     0.40 &         0.65\\
50\%   &     0.00 &     0.90 &     0.00 &     0.00 &     0.57 &     0.85 &     0.61 &     0.59 &     0.53 &     0.51 &     0.53  &         0.67\\
75\%   &     0.01 &     0.91 &     0.00 &     0.00 &     0.66 &     0.88 &     0.71 &     0.71 &     0.74 &     0.72 &     0.72  &         0.68 \\
Max   &     1.00 &     0.99 &     1.00 &     1.00 &     1.00 &     1.00 &     1.00 &     1.00 &     1.00 &     1.00 &     1.00 &         0.96 \\

\bottomrule
\end{tabular}
\end{table}

\begin{figure}[h!]
\centering
\includegraphics[width =0.8\textwidth]{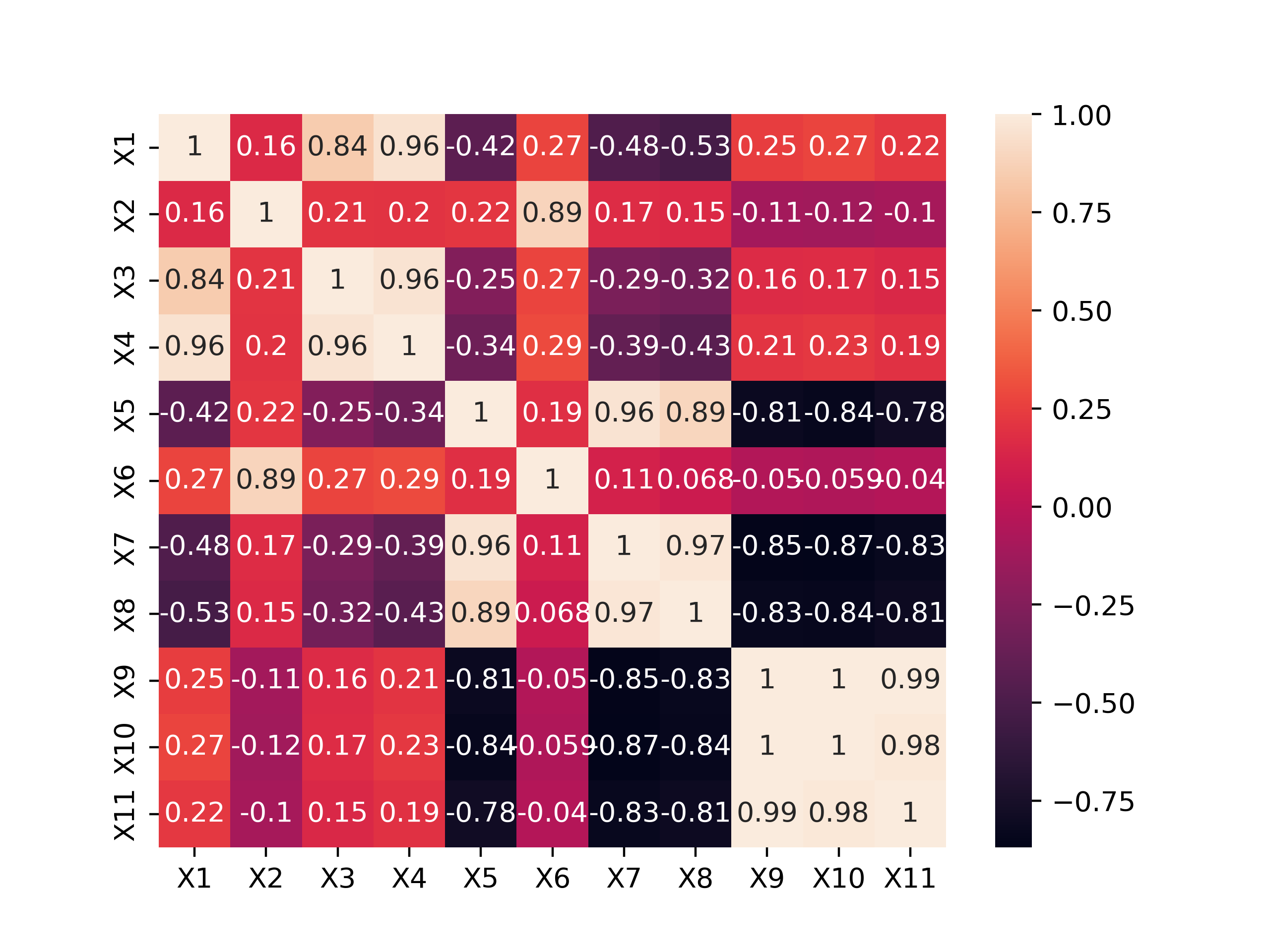}
\caption{Correlation Heatmap of Numerical Features (excluding variable Y).}
\label{heat_map}
\end{figure}

Correlation heatmap between numerical features (excluding variable Y) is provided in Figure \ref{heat_map}. We observed the feature $X3$ and the feature $X4$ are highly positively correlated. In addition, the feature $X7$ and $X8$ have a highly positive correlation. Furthermore, the feature $X9$, $X10$ and $X11$ are highly positively correlated.

\subsection{Methods and implementation}

\subsubsection{Candidate Methodologies} 

A few models were employed to test the proposed framework in predicting the water quality.

\paragraph{\textbf{Benchmark}}

A benchmarking approach is commonly required to rationalize the necessity of employing advanced models. In this study, we take the arithmetic average value of the target variable Y in the training set as a prediction as a benchmark. 

\paragraph{\textbf{Spatial-Autoregressive Dependency Learning II (SADL-II) }}

A spatio-temporal model (SADL-II) was proposed in the original paper \cite{zhao2019spatial}. On the basis of spatial topological restrictions, SADL-II was suggested to automatically learn the conditional independence pattern as well as quantify the numerical values of the spatial dependency. It is considered as an alternative benchmarking model in this paper.

\paragraph{\textbf{Linear Regression with Elastic Net Regularization}}

Multiple Linear Regression is a statistical technique used to model the relationship between multiple independent variables and a dependent variable. It extends the concept of simple linear regression by considering multiple predictors. The general form of Multiple Linear Regression is as follows:
\begin{equation}
Y = \beta_0+\beta_1 \times X_1+\cdots + \beta_n \times X_n +\epsilon,
\end{equation}

where $\beta_0, \beta_1, \cdots \beta_n$ are $n$ coefficients and $\epsilon$ is a random error, $Y$ is the target variable, and $X_1, X_2,\cdots, X_n$ are predictors.

Elastic Net is a linear regression technique that combines the features of both L1 (Lasso) 
 \cite{tibshirani1996regression} and L2 (Ridge) \cite{hoerl1970ridge} regularization methods. It is a regularization and variable selection technique used in machine learning and statistics to prevent overfitting and improve the stability and generalization of linear regression models. The loss function is as follows:
\begin{equation}
\mathcal{L} = \frac{\sum^n_{i=1}(y_i-x'_i\beta)^2}{2n} + \lambda\frac{1-\alpha}{2}\sum^k_{j=1}\beta^2_j+\lambda\alpha|\beta_j|.
\end{equation}
In this loss function, the new parameter $\alpha$ is a ``mixing" parameter that balances the two approaches. If $\alpha$ is one, this is Lasso regression, and if $\alpha$ is zero, then this is Ridge regression.

Lasso can improve prediction accuracy and model interpretability by performing both feature selection and regularization. Lasso's feature selection property provides a clear interpretation of the most important predictors in the model. The non-zero coefficients indicate the features that strongly impact the prediction outcome \cite{tibshirani1996regression}. 

\paragraph{\textbf{XGBoost}}
XGBoost was first introduced in \cite{chen2016xgboost}. They provided an in-depth explanation of the algorithm, its optimization techniques, and empirical evaluation of its performance on various datasets. XGBoost scales beyond billions of examples using far fewer resources than existing systems. The  objective function (loss function and regularization) at iteration $t$ that we need to minimize is the following:

\begin{equation}
\mathcal{L}^{(t)} = \sum^n_{i=1}l(y_i,\hat{y}_i^{(t-1)}+f_t(x_i)) + \Omega(f_t),
\end{equation}
where $l$ is the loss function, $f_t$ is the $t-$th tree output, and $\Omega$ is the regularization term to control the complexity of the model and prevent the model from overfitting.

\paragraph{\textbf{LightGBM}}
LightGBM, an open-source machine learning framework created by Microsoft, is an optimized tool tailored for gradient boosting. Gradient boosting is a widely-used machine learning approach that constructs predictive models by combining the outcomes of numerous weak models, often in the form of decision trees. One of the primary distinctions that sets LightGBM apart from the traditional gradient boosting tree decision technique is the use of a method called GOSS (Gradient-based One-Side Sampling).

During the training process, GOSS retains all the data points with significant gradients while randomly subsampling the data with lower gradients. This strategic approach effectively reduces the search space, enabling GOSS to converge more swiftly.

LightGBM has gained acclaim for its remarkable speed and efficiency, rendering it a preferred choice across a range of machine learning tasks, including classification, regression, and ranking.

\paragraph{\textbf{Random Forest}}
In 2001, Leo Breiman \cite{breiman2001random} presented the Random Forest algorithm and discussed its principles and advantages. Random Forest is a powerful ensemble learning method in machine learning, primarily used for classification and regression tasks. It is an ensemble of decision trees, where multiple decision trees are trained independently, and their outputs are combined to make predictions. 

\paragraph{\textbf{Multiple-layer Perceptron (MLP) neural network}}
Multiple-layer Perceptron (MLP) neural network represents an artificial neural network model that leverages the backpropagation technique to iteratively refine the connections between neurons, thereby enhancing its predictive accuracy. This implementation integrates the Multi-Layer Perceptron (MLP) algorithm \cite{lecun1998gradient}, harnessing backpropagation and stochastic gradient descent strategies for training and evaluating datasets. It offers a range of customizable parameters, allowing users to finely tune the model's performance by adjusting factors such as the number of hidden layers, activation functions, optimization solvers, and more. MLPRegressor stands as an effective solution for addressing regression tasks, demonstrating proficiency in capturing intricate non-linear relationships between input and output variables.

\subsection{Implementation}
In this section, we detail the proposed prediction framework through hyperparameter tuning, generating predictions, and assessing their effectiveness using tailored scoring metrics. Additionally, we capture the time taken for these tasks and endeavor to identify disparities in feature selection, as evidenced by variations in feature importance.

\begin{enumerate}
    \item Choose the Scoring Function: For all candidate methods, we use the negative root of mean squared error as the scoring function for hyperparameter tuning and evaluation.

    \item Define Hyperparameters: Set up the hyperparameter space for each method.

    \item Start Time Recording: Begin recording the time taken for the entire procedure.

    \item Initialize Models: Initialize the models for each candidate method.

    \item Hyperparameter Tuning: Perform hyperparameter tuning using cross-validation to find the best hyperparameters for each method.

    \item Final Predictions: Make final predictions on the testing data using the best-tuned models.

    \item Score Calculation: Calculate the evaluation metrics on the test data for each method.

    \item End Time Recording: Stop recording the time taken for the procedure.

    \item Feature Importance: Obtain feature importance scores from the best-tuned models.

    \item Save Results: Save the results of each method's performance and feature importance for model comparison.
\end{enumerate}


\begin{figure}[h]
\begin{center}
\includegraphics[width=6cm, height=10cm]{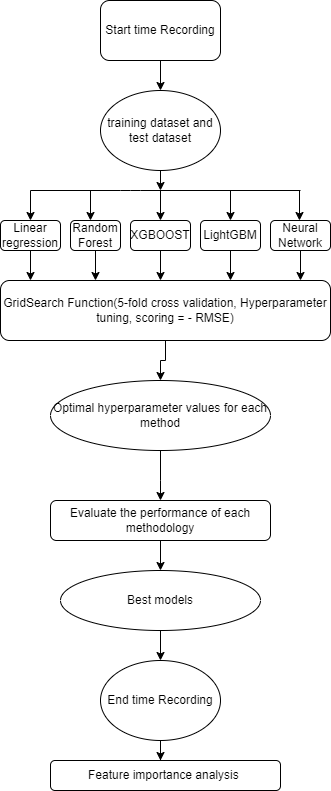}
\caption{Flowchart of The Proposed Water Quality Prediction Framework.}
\end{center}
\end{figure}

\subsection{Model Selection}
\subsubsection{Evaluation Metrics}

Based on the nature of the problem, and to compare against the original paper, we have chosen the following metrics: root mean square error (RMSE), mean absolute percentage error (MAPE), weighted mean absolute percentage error (WMAPE), weighted under prediction (WUPRED) and weighted over prediction (WOPRED). These metrics are defined by the following equations:

\begin{equation}
RMSE =  \sqrt{\sum^n_{i=1}\frac{(y_i - \hat{y_i})^2}{n}},
\end{equation}


\begin{equation}
MAPE = \frac{1}{n}\sum^n_{i=1}|\frac{y_i-\hat{y}_i}{y_i}|,
\end{equation}

\begin{equation}
WMAPE = \frac{\sum^n_{i=1}|y_i-\hat{y}_i|}{\sum^n_{i=1}|y_i|},
\end{equation}


\begin{equation}
WUPRED = \frac{\sum^n_{i=1} I\{y_i > \hat{y}_i\}(y_i-\hat{y}_i)}{\sum^n_{i=1}y_i},
\end{equation}

\begin{equation}
WOPRED
 = \frac{\sum^n_{i=1} I\{y_i < \hat{y}_i\}(\hat{y}_i - y_i)}{\sum^n_{i=1}y_i},
\end{equation} 

where $y_i$ represents the observed PH value, $\hat{y}_i$ is the value of prediction. In the context of all five metrics, our goal is to minimize them as much as possible. RMSE assesses the error magnitude in the same units as the predicted values. In contrast, MAPE highlights relative percentage errors, effectively mitigating the influence of outliers on evaluations. WMAPE builds upon MAPE by introducing weighted adjustments that account for the varying importance of different data points. Meanwhile, WUPRED and WOPRED quantify errors when forecasts fall short or exceed actual values, shedding light on potential model biases. Collectively, these metrics provide a comprehensive evaluation of predictive performance, considering both error direction and magnitude, and can be invaluable in guiding enhancements to predictive models.

\subsubsection{Results}
This section presents the prediction results and error analysis results of PH values. During the cross validation process, $5$-fold cross validation is selected. To test if the spatio-temporal features are contributing to the predictions, three model implementation strategies are derived:

\begin{enumerate}
    \item \textbf{Strategy 1}: We select eleven numerical features as input features to predict the target variable.
    \item \textbf{Strategy 2}: We select the standardized eleven numerical features as input features to predict the target variable.
    \item \textbf{Strategy 3}: The standardized numerical features and the one-hot encoded categorical features are selected to predict the target variable.
\end{enumerate}

For strategy 1, the model performance summary is shown in Table \ref{prediction_numerical_only}. Based on the performance of Benchmarking and other advanced models, we can conclude that employing advanced models is essential. In addition, the XGBoost beats all the other candidate models in terms of MAPE, WMAPE and WOFOREC. For metric WUFOREC, MLP has the best performance. Meanwhile, the lightGBM outperforms the other models under the metric RMSE. Tree-based methods have shown their superiority in regression problems. There are several candidate models outperform the SADL-II (Original Paper).  

\begin{table}[h]
\caption{Water Quality Prediction Results (Numerical Features Only).}
\label{prediction_numerical_only}
\begin{tabular}{@{}lrrrrr@{}}
\toprule
    Models  &     RMSE      & MAPE & WMAPE  & WUPRED & WOPRED \\ 
    \midrule

SADL-II (Original Paper) & 11.50 & N/A  & N/A  & N/A  & N/A  \\

Benchmarking                  & 29.52 & 32.21 &  32.35  &    14.55 &    17.80 \\
LightGBM & \textcolor{red}{10.88} &  9.34 &   9.48 &        5.20 &     4.28 \\
Linear Regression   & 12.04 & 10.72 &  10.90 &       5.65 &     5.24 \\
MLP     & 11.62 & 10.00 &  10.15 &        \textcolor{red}{5.13} &     5.02 \\
Random Forest       & 11.47 &  9.71 &   9.87 &        5.34 &     4.52 \\
XGBoost     & 10.96 &  \textcolor{red}{9.23} &   \textcolor{red}{9.37} &       5.17 &     \textcolor{red}{4.19} \\

\bottomrule
\end{tabular}
\footnotetext{\emph{Note.} The numbers in \textcolor{red}{red} color stand for the best performance under the specific metric. All values are their original values $\times$ 1000.}
\end{table}%

For strategy 2 (model performance summary detailed in Table \ref{prediction_numerical_only_std}), we have the same conclusion for the tree based models as what we had in strategy 1. In addition, the performance of XGBoost and LightGBM had a slight improvement after the standardization of numerical features. However, the performance of MLP got worse than the linear regression with elastic net regularization after the standardization.  

\begin{table}[h]
\caption{Water Quality Prediction Results (Numerical Features Only; Standardized).}
\label{prediction_numerical_only_std}
\begin{tabular}{@{}lrrrrr@{}}
\toprule
    Models  &     RMSE      & MAPE & WMAPE & WUPRED & WOPRED \\ 
\midrule
SADL-II (Original Paper) & 11.50 & N/A  & N/A  & N/A  & N/A  \\
Benchmarking                  & 29.52 & 32.21 &  32.35 &      14.55 &    17.80 \\
LightGBM & \textcolor{red}{10.84} &  9.33 &   9.46 &       \textcolor{red}{5.16} &     4.30 \\
Linear Regression    & 12.04 & 10.74 &  10.92 &       5.66 &     5.25 \\
MLP     & 14.20 & 12.92 &  13.16 &       3.83 &     9.33 \\
Random Forest      & 11.47 &  9.71 &   9.87 &        5.35 &     4.51 \\
XGBoost      & 10.96 &  \textcolor{red}{9.22} &   \textcolor{red}{9.36} &        5.17 &     \textcolor{red}{4.19} \\
\bottomrule
\end{tabular}
\footnotetext{\emph{Note.} The numbers in \textcolor{red}{red} color stand for the best performance under the specific metric. All values are their original values $\times$ 1000.}
\end{table}%

In strategy 3 (model performance detailed in Table \ref{prediction_numerical_Categorical}), the spatio-temporal features are involved. The lightGBM beats all the other models when it comes to the metric RMSE, MAPE and WMAPE. However, the general performance of models slightly degraded after the spatio-temporal features are inputted. Same as the results in strategy 1 and strategy 2, there are several candidate models outperform the SADL-II (Original Paper).

\begin{table}[h]
\caption{Water Quality Prediction Results (Numerical Features: Standardized; Categorical Features: One-Hot Encoding).}
\label{prediction_numerical_Categorical}
\begin{tabular}{@{}lrrrrr@{}}
    \toprule
    Models  &     RMSE      & MAPE & WMAPE & WUPRED & WOPRED \\ 
    \midrule
SADL-II (Original Paper) & 11.50 & N/A  & N/A  & N/A  & N/A  \\
Benchmarking                  & 29.52 & 32.21 &  32.35 &      14.55 &    17.80 \\
LightGBM & \textcolor{red}{11.04} &  \textcolor{red}{9.22} &   \textcolor{red}{9.36} &      \textcolor{red}{5.15} &     4.21 \\
Linear Regression     & 11.91 & 10.79 &  10.96 &      6.14 &     4.82 \\
MLP      & 14.42 & 14.54 &  14.61 &      9.29 &     5.32 \\
Random Forest       & 11.71 &  9.71 &   9.87 &       5.26 &     4.61 \\
XGBoost     & 11.14 &  9.29 &   9.44 &        5.28 &     \textcolor{red}{4.16} \\
\bottomrule
\end{tabular}
\footnotetext{\emph{Note.} The numbers in \textcolor{red}{red} color stand for the best performance under the specific metric. All values are their original values $\times$ 1000.}
\end{table}%

We also provided hyper-parameters tuning time for strategy 2 and strategy 3 in Table \ref{Time_numerical_only_std} and Table \ref{Time_prediction_all}. In the tables, ``Total Fits" stands for the number of fits in hyper-parameters tuning of each model. ``Tuning Time" is for running time of entire hyper-parameters tuning process of each model. ``Fitting Time (Best Model)" stands for the training time for the selected best hyper-parameters of each model. ``Average Tuning" is for the average hyper-parameters tuning time. The tuning time for strategy 1 is similar to the one for strategy 2, therefore, it is not provided in the paper. In addition, the benchmark model does not have a hyper-parameters tuning process. In conclusion, the lightGBM has a lowest average tuning time when the spatio-temporal features are included. Otherwise, he linear regression with elastic net regularization is the most efficient one.

\begin{table}[h]
\caption{Running Time Summary (Numerical Features: Standardized).}
\label{Time_numerical_only_std}
\begin{tabular}{@{}lrrrr@{}}
\toprule
{Models} &  Total Fits &  Tuning Time  &  Average Tuning &  Fitting Time (Best Model)\\
    \midrule
LightGBM &            2000 &          159.92  &            0.08 &                     0.19 \\
Linear Regression      &             150 &            3.32  &            0.02 &                     0.07 \\
MLP      &              40 &         1948.63  &           48.72 &                   116.05\\
Random Forest      &            1200 &         1934.51 &            1.61  &                    64.58 \\
XGBoost     &            5760 &        2429.88 &            0.42 &                     4.30 \\
\bottomrule
  \end{tabular}
\footnotetext{\emph{Note.} The unit for running time is second.}
\end{table}%

\begin{table}[h]
\caption{Running Time Summary in Seconds (Numerical Features: Standardized; Categorical Features: One-hot encoding).}
\label{Time_prediction_all}
\begin{tabular}{@{}lrrrr@{}}
\toprule
{Models} &  Total Fits &  Tuning Time&  Average Tuning  &  Fitting Time (Best Model) \\
    \midrule
LightGBM &            2000 &       213.44  &            0.11 &                     0.46\\
Linear Regression     &             150 &        44.39  &            0.30 &                     0.82\\
MLP     &              40 &      2147.05  &           53.68 &                    65.81\\
Random Forest       &            1200 &      3633.05  &            3.03 &                   115.67 \\
XGBoost      &            5760 &     13320.13  &            2.31 &                     4.16 \\

\bottomrule
  \end{tabular}
\footnotetext{\emph{Note.} The unit for ruining time is second.}
\end{table}%

\subsection{SHapley Additive exPlanations (SHAP)}
For each prediction, the SHAP values associated with environmental variables serve to quantify their localized contributions to that specific prediction 
 \cite{NIPS2017_8a20a862}. The mathematical definition of the SHAP value is provided below:

\begin{align*}
\Phi_i(F, x) &= \sum_{Z'\subseteq X'} [ |Z'|\times \left(\begin{array}{l}M \\ |Z'|\end{array}\right)]^{-1}[F(Z')-F(Z'\backslash x_i)] \\
&=\sum_{Z'\subseteq X'}\frac{(|Z'|-1)!(M-|Z'|)!}{M!}[F(Z')-F(Z'\backslash x_i)].
\end{align*}

In this equation, $\Phi_i(F, x)$ represents the SHAP value corresponding to feature $x_i$ in the context of a model $F$ constructed on a set of features $X$. $M$ represents the total number of input features, $X'$ denotes the set of all potential feature combinations that include feature $x_i$, and $|Z'|$ signifies the number of features within a particular feature combination $|Z'|$. Additionally, $F(Z')$ and $F(Z'\backslash x_i)$ represent distinct predictive models trained on $|Z'|$ and $Z'\backslash x_i$ (which is $Z'$ with feature $x_i$ removed), respectively. Thus, the SHAP value is determined by aggregating the marginal contributions $F(Z')-F(Z'\backslash x_i)$ from all feasible feature combinations $(Z')$ through a weighted average.

SHAP values are computed using the Python implementation of SHAP, as described by Lundberg and Lee in 
 \cite{NIPS2017_8a20a862}. In this study, we calculated SHAP values for some candidate models for all strategies. They provided similar results crossing the strategies, so we diplayed the visualization of SHAP values for strategy 2 for illustration purposes. 

\begin{figure}[H]		\centering
        \includegraphics[width=0.9\linewidth]{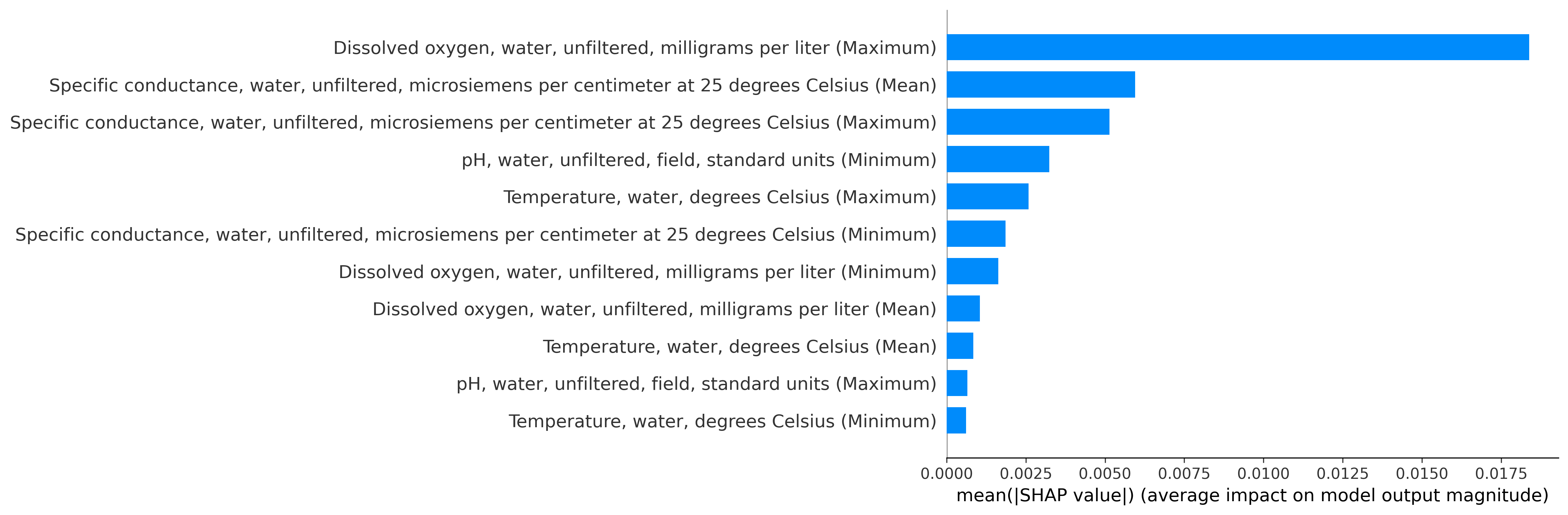}
    \caption{Average Impact (SHAP value) on model output for LightGBM. }\label{best_lightGBM_shap_bar}
\end{figure}

\begin{figure}[H]		\centering
        \includegraphics[width=0.9\linewidth]{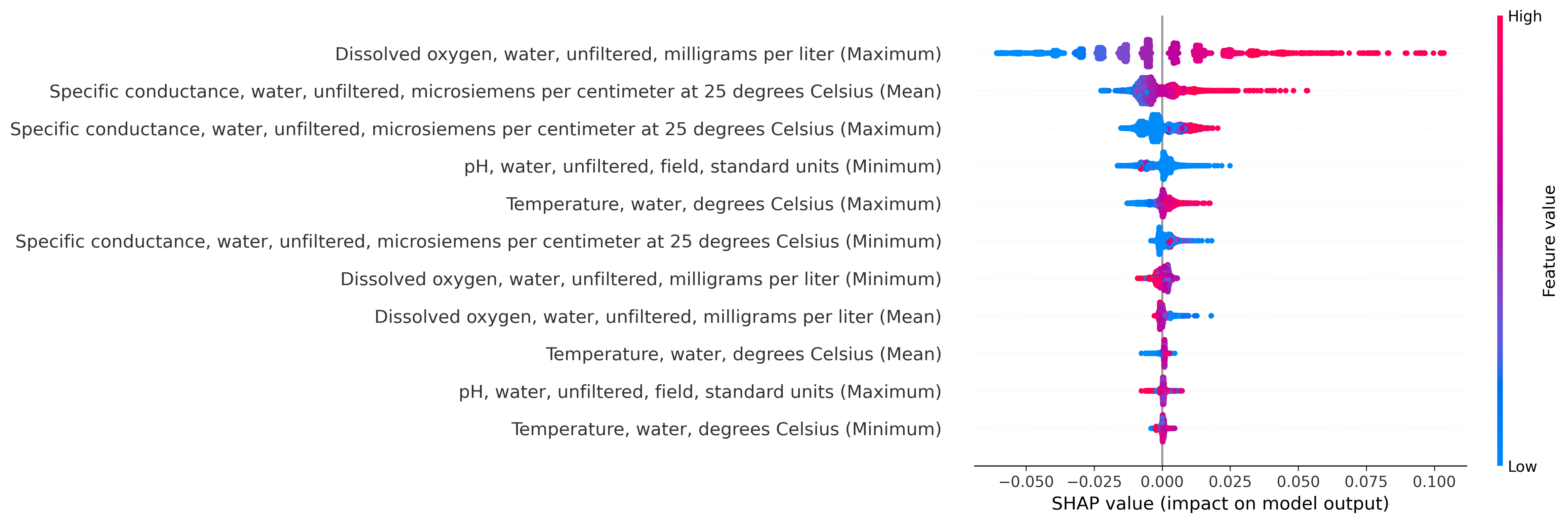}
    \caption{Impact (SHAP value) on model output for LightGBM.}\label{best_lightGBM_shap_val}
\end{figure}

\begin{figure}[H]		\centering
        \includegraphics[width=0.9\linewidth]{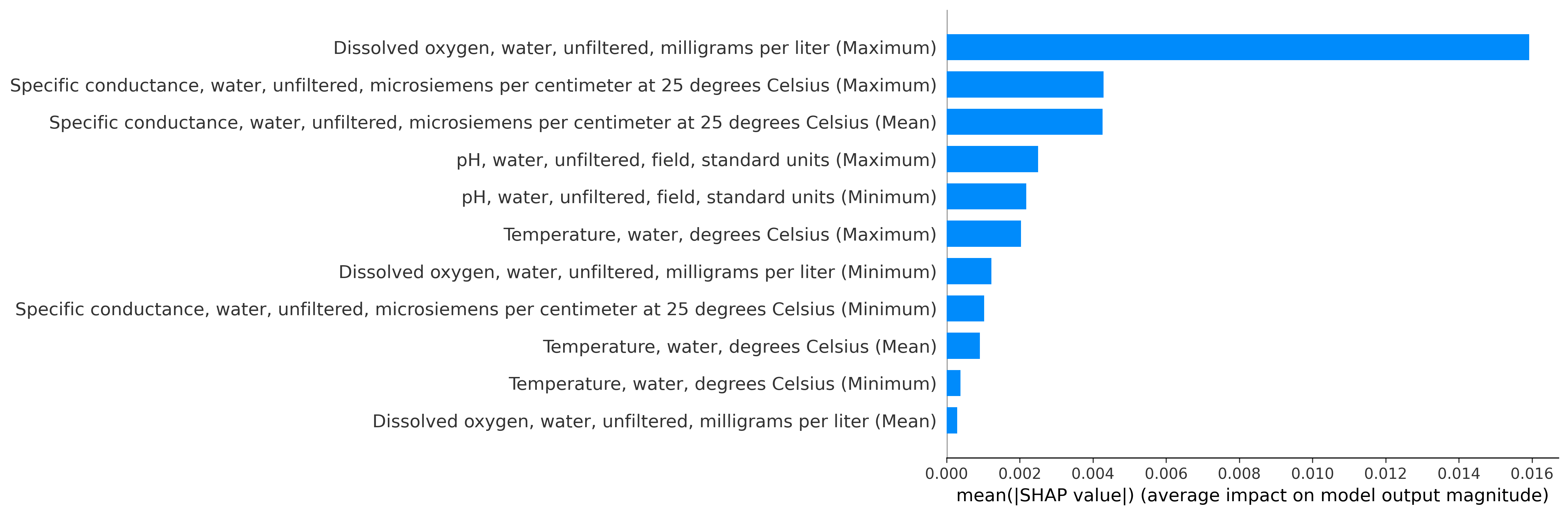}
    \caption{Average Impact (SHAP value) on model output for  XGBoost.}\label{best_xgb_shap_bar}
\end{figure}

Figure \ref{best_lightGBM_shap_bar} and Figure \ref{best_lightGBM_shap_val} are the illustration of SHAP values for lightGBM model. From the plots, the feature X6 (``Dissolved oxygen, water, unfiltered, milligrams per liter (Maximum)") has the significant impact in predicting the target variable. It's impact is much higher than the other features. From Figure \ref{best_xgb_shap_bar}, XGBoost's SHAP values generally agree with the SHAP values of lightGBM.

\begin{figure}[H]		\centering
        \includegraphics[width=0.9\linewidth]{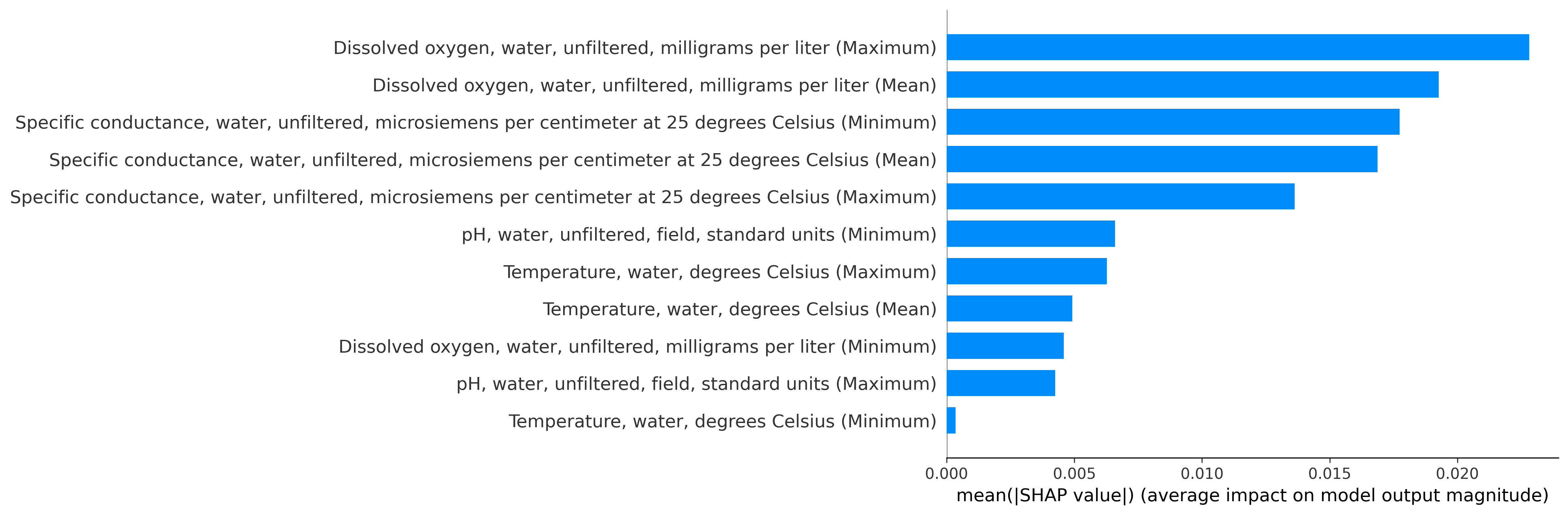}
    \caption{Average Impact (SHAP value) on model output for Linear Regression.}\label{best_lin_shap_bar}
\end{figure}


\section{Findings}

We found some machine learning models are showing better performance than the spatial topological model described in the original paper. There could be several reasons why the machine learning models outperform the SADL-II model in the original paper that takes into account time dependency and spatial features. Here are a few possible explanations:

\subsection{Data Quality and Quantity} The availability of a larger and higher-quality dataset can have a significant impact on model performance. The data provided by the original paper is extensive and contains no missing values, the machine learning models benefit from having more reliable information.

\subsection{Feature} The efficacy of machine learning models is profoundly influenced by the quality and pertinence of the features employed for training. It is conceivable that the features provided by the original paper and we meticulously engineered have proven to be insightful and pertinent in predicting water pH values. Remarkably, the machine learning approaches leverage the automated feature selection during training time, which is done via the feature importance scores, aiding in feature selection and engineering.


\subsection{Model Complexity and Flexibility} Machine learning models, especially ensemble methods like Random Forest and XGBoost, have the capability to capture complex relationships and patterns in the data. They can automatically learn interactions between features and nonlinear relationships that the original model might not have been able to capture effectively.

These ensemble models combine multiple individual models to make predictions. This aggregation reduces the risk of overfitting, as the errors of individual models tend to cancel out, leading to more robust predictions on unseen data. And they strike a balance between bias and variance. By averaging or combining predictions from multiple models, they reduce variance while maintaining a low bias, resulting in more accurate predictions.

\subsection{Hyperparameter Tuning}

We employ a systematic hyperparameter tuning process for all candidate models. This process aims to strike a balance among model complexity, interpretability, and predictability. It enables us to find the model that best fits the data, even without prior expert knowledge in water quality field.



\section{Discussion}

In our investigation to predict water pH values, we adopted several Machine learning approaches and extended our analysis to include temporal and spatial features. Surprisingly, despite adding these features, the improvement in prediction accuracy was not as significant as expected. However, the results still demonstrated superior performance compared to the original temporal spatial model proposed by Chang et.al 
 \cite{chang2016estimating} in their study.

Our findings challenge the belief (from the original paper) that incorporating time-dependent and spatial features would invariably enhance the accuracy of predictive models. The success of machine learning models lies in the meticulous process of feature selection, engineering, and model optimization.

Suitable feature engineering helps the machine models find hidden patterns in the data that the original model didn't catch. Also, ensemble tree methods like LightGBM and XGBoost are good at finding complicated patterns in the data.


In comparison to the original model's approach of leveraging time dependencies and spatial dependencies, the machine learning models exhibited higher predictive accuracy. This discrepancy may be attributed to the more sophisticated modeling techniques and the capacity to harness the power of the ensemble approach, both of which are integral to modern machine learning

Our study underscores the importance of a holistic approach to predictive modeling, wherein data preprocessing, feature engineering, model selection, and hyperparameter tuning collectively contribute to the overall performance. While time-dependent and spatial features can certainly enhance predictive capabilities, our results emphasize the critical role of model selection and optimization in achieving remarkable accuracy.

In summary, our exploration into predicting water pH values with machine learning has not only yielded superior results compared to the original model but also shed light on the multifaceted nature of predictive modeling. As the field of machine learning continues to evolve, the intricate interplay between features, algorithms, and optimization techniques emerges as a pivotal factor in determining the success of predictive endeavors. We will test this standardized framework on other spatial-temporal prediction tasks in future work.

\paragraph{Acknowledgments}
The source code for this project is available on GitHub at the following URL: \href{https://github.com/YinpuLi/water-quality-prediction.git}{water-quality-prediction}. The repository contains the implementation of the algorithms and methods comparison discussed in this paper.

\bibliography{sn-bibliography}

\end{document}